\documentclass[journal]{IEEEtran}

\usepackage{cite}

\ifCLASSINFOpdf
\else
\fi
\usepackage{amsmath}

\usepackage{algorithm}
\usepackage{algorithmic}
\usepackage{balance}

\usepackage{array}
\ifCLASSOPTIONcompsoc
  \usepackage[caption=false,font=normalsize,labelfont=sf,textfont=sf]{subfig}
\else
  \usepackage[caption=false,font=footnotesize]{subfig}
\fi

\usepackage{graphicx}

\usepackage{url}
\usepackage{color}

\begin{document}
%\singlespacing
\title{Coarse-to-fine Seam Estimation for Image Stitching}

\author{Tianli~Liao\thanks{T. Liao is with the Center for Combinatorics, Nankai University, Tianjin 30071, China. Email: liaotianli@mail.nankai.edu.cn}, Jing Chen\thanks{J. Chen is with the Center for Combinatorics, Nankai University, Tianjin 30071, China. Email: chenjing@mail.nankai.edu.cn} and Yifang Xu\thanks{Y. Xu is with the Center for Combinatorics, Nankai University, Tianjin 30071, China. Email: xyf@mail.nankai.edu.cn}}

% <-this % stops a space


\maketitle

\begin{abstract}
    Seam-cutting and seam-driven techniques have been proven effective for handling imperfect image series in image stitching. Generally, seam-driven is to utilize seam-cutting to find a best seam from one or finite alignment hypotheses based on a predefined seam quality metric. However, the quality metrics in most methods are defined to measure the average performance of the pixels on the seam without considering the relevance and variance among them. This may cause that the seam with the minimal measure is not optimal (perception-inconsistent) in human perception. In this paper, we propose a novel coarse-to-fine seam estimation method which applies the evaluation in a different way. For pixels on the seam, we develop a patch-point evaluation algorithm concentrating more on the correlation and variation of them. The evaluations are then used to recalculate the difference map of the overlapping region and reestimate a stitching seam. This evaluation-reestimation procedure iterates until the current seam changes negligibly comparing with the previous seams. Experiments show that our proposed method can finally find a nearly perception-consistent seam after several iterations, which outperforms the conventional seam-cutting and other seam-driven methods.
\end{abstract}


% \fi
\IEEEpeerreviewmaketitle

\section{Introduction}

\IEEEPARstart{I}{mage} stitching for imperfect image series is a challenging problem which has gained much progress in recent years \cite{szeliski2006image,Brown:2007}. Generally, there are two ways to address this problem. One way is to propose an alignment technique (image warping) that aligns the images as accurate as possible \cite{gao2011constructing,lin2011smoothly,zaragoza2014projective,chen2016natural,zhang2016multi}. Another way called seam-driven (or seam-guided) is to utilize the seam-cutting \cite{kwatra2003graphcut,agarwala2004interactive} to find a most invisible seam in the overlapping region from one or finite alignment hypotheses \cite{gao2013seam,zhang2014parallax,lin2016seagull,li2018perception}. The first way aims to generate a geometrically accurate result, which may fail to be effective when the input images have parallax or other issues. Thus seam-driven becomes the critical way to produce convincing stitching results.

\begin{figure}
	\centering
	\subfloat[Seam with the quality measure equals 0.140.]{
	\includegraphics[width=0.23\textwidth]{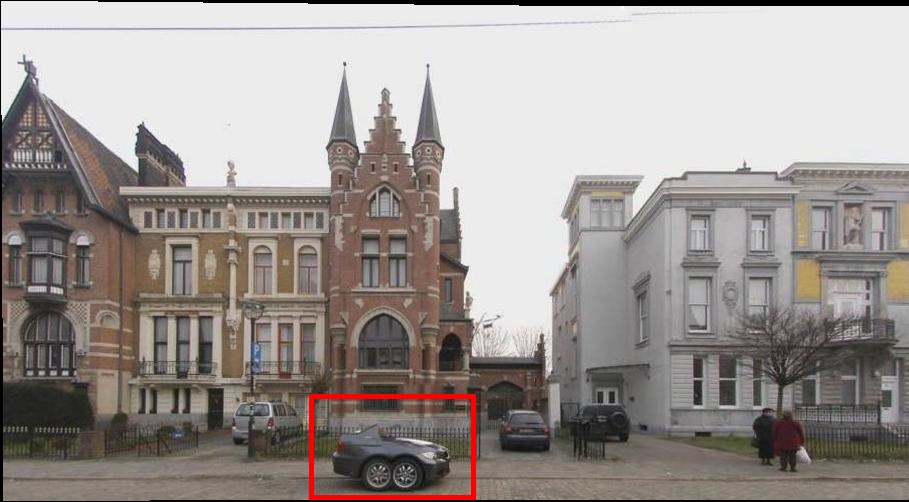}
	\includegraphics[width=0.23\textwidth]{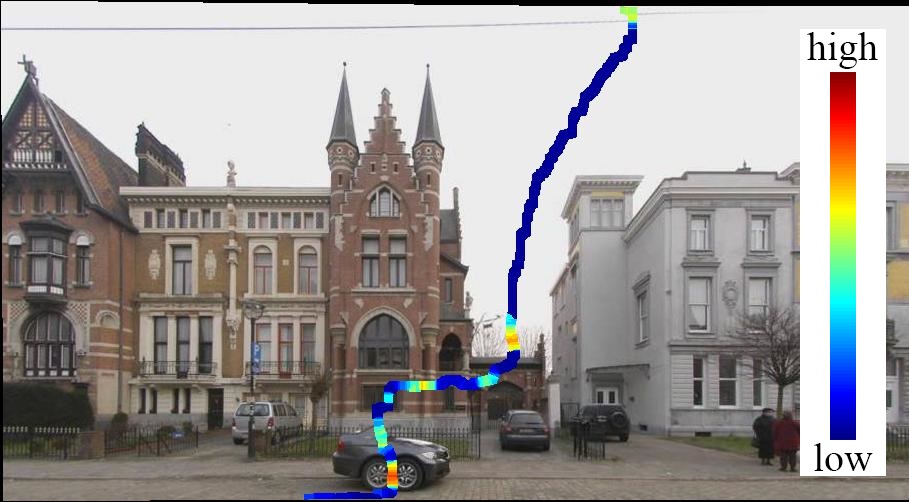}}\\
	\subfloat[Seam with the quality measure equals 0.165.]{
	\includegraphics[width=0.23\textwidth]{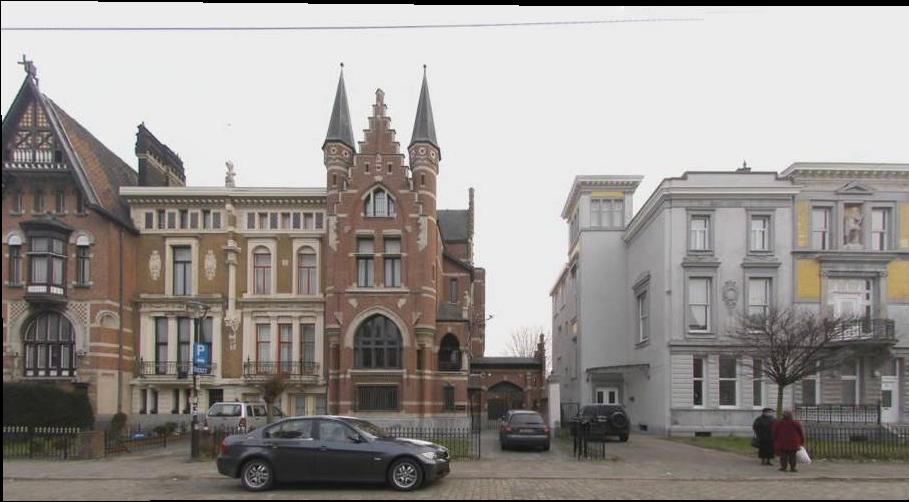}
	\includegraphics[width=0.23\textwidth]{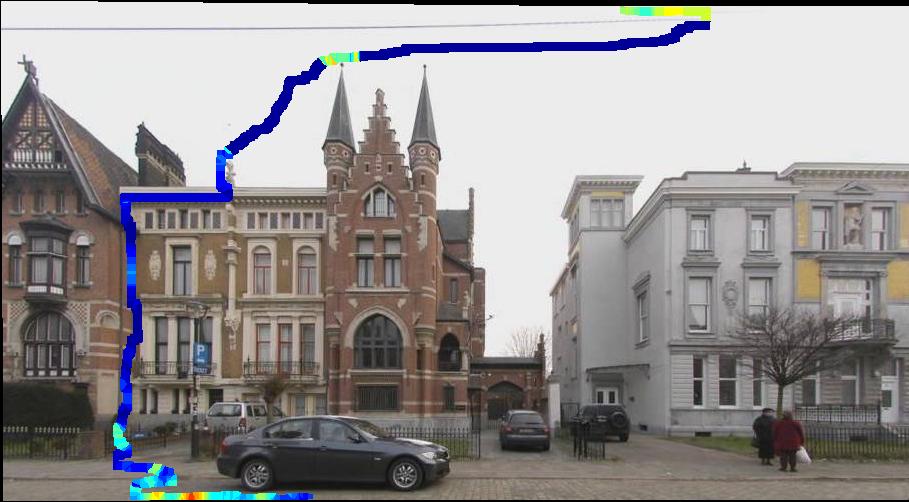}}\\
	\caption{Result comparison between two stitching seams where the seams are shown as a hot map based on the quality metric defined in \cite{lin2016seagull}. The input images are from \cite{zhang2014parallax}. (a) The final result and the corresponding seam with a smaller measure. (b) The final result and the corresponding seam with a bigger measure.}
	\label{fig:1}
\end{figure}

The seam-driven strategy for image stitching is first proposed by Gao \emph{et al.} \cite{gao2013seam}. They applied seam-cutting to estimate multiple seams from a finite alignment hypotheses. Then a seam quality metric is defined to evaluate these seams and the final result is produced from the seam with the minimal measure. This pipeline is adopted in many other seam-driven methods \cite{zhang2014parallax,lin2016seagull}. However, their quality metrics are defined to measure the average performance of the pixels on the seam without considering the relevance and variance among them. This may cause that the seam with the minimal measure is not optimal in human perception. Fig. \ref{fig:1} shows a comparison example where two seams are measured. It is worth noting that there are some inaccurate measurements (false positives) for the pixels on the seam. In fact, it is difficult to define a single quality metric to precisely evaluate the stitching seams, since in the seam-driven strategy, two seams can be equally convincing in human perception despite their distinct quality metric. This motivates us to develop a seam estimation method which can find the perceptually optimal seam given one alignment hypothesis.
%In fact, it is objectively difficult to define a single metric to evaluate the stitching seams, since in a seam-driven strategy, the goal is to produce a seamless result in human perception instead of providing a geometrically accurate one \cite{junhong2013new}. 

In this paper, we propose a coarse-to-fine seam estimation method for image stitching. From the perspective of seam evaluation, we observe that a perceptually optimal seam should have a relatively small quality measure as well as a small variance of the pixels on the seam (see Fig. \ref{fig:refine}(d)). Our coarse-to-fine strategy has two main steps. In the first step, given aligned images, we estimate a stitching seam via the conventional seam-cutting where the energy function is calculated based on the original difference map. In the second step, we introduce a patch-point evaluation algorithm to evaluate the pixels on the seam, the evaluations are then used to recalculate the difference map and reestimate a stitching seam. The two processes iterate until the current seam changes negligibly comparing with the previous seams. Experiments show that our method outperforms the conventional seam-cutting and other seam-driven methods. %{\red Besides, it roughly points out a way to objectively evaluate the stitching seam.}

\section{Related Work}

In recent years, many efforts have been devoted to seam-cutting or seam-driven to address the complex scenes and issues in image stitching. Seam-cutting is proposed to handle the imperfect image series which aims to estimate an invisible seam between overlapping images such that the images can be seamlessly blended together. Most seam-cutting methods formulate the seam estimation as the energy minimization of a labeling problem, and minimize the energy via graph cuts \cite{boykov2001fast}. Various energy functions were defined in their work to handle specific issues \cite{kwatra2003graphcut,agarwala2004interactive,Eden2006seamless,jia2008image,zhang2016multi,li2018perception}. Our method takes the conventional seam-cutting as the initial seam estimation method.

%The seam-cutting approach is then incorporated into the seam-driven framework.
The seam-driven methods then incorporate the seam-cutting approaches in their framework.
Gao \emph{et al.} \cite{gao2013seam} indicated that the perceptually best result is not necessarily from the best global alignment. To find the best result from multiple seams, they defined a seam quality metric to measure the stitching seams.
Zhang and Liu \cite{zhang2014parallax} improved this strategy by combing homogarphy and content-preserving warps \cite{liu2009content} to locally align images and generate better alignment hypotheses, where the seam cost is used as the quality metric. Lin \emph{et al.} \cite{lin2016seagull} proposed to generate the alignment hypotheses via a superpixel-based feature grouping and a seam-guided structure-preserving warp, where the warp is iteratively improved by adaptive feature weighting. They also defined a quality metric based on the ZNCC (zero-mean normalized cross correlation) score, which was also used in \cite{li2018perception}. All these quality metrics in seam-driven are defined to evaluate the average performance of the pixels on the seam, it may cause that the seam with the minimal measure is not optimal in human perception.

Our method adopts the seam evaluation in a different way. Instead of defining a seam quality metric to find the seam of best performance from multiple seams, we propose a evaluation algorithm concentrating more on the correlation and variation of the pixels on the seam. The evaluation is then applied into our coarse-to-fine seam estimation strategy.

\section{Coarse-to-fine Seam Estimation}

In this section, we first give a brief description about the conventional seam-cutting method and propose our patch-point evaluation algorithm, including a patch evaluation and a point evaluation. Then, we develop an iterative evaluation-reestimation procedure and summarize our coarse-to-fine seam estimation framework in the end.

\subsection{Conventional Seam-cutting}

For two-image stitching, we use $I_0$ and $I_1$ to denote the aligned reference and target images, $\mathcal{P}$ is their overlapping region and $\mathcal{L}=\{0,1\}$ is a label set. A seam means assigning a label $l_p\in \mathcal{L}$ to each pixel $p\in \mathcal{P}$ where ``0'' corresponds to $I_0$ and ``1'' corresponds to $I_1$. Seam-cutting aims to find a labeling $l$ (i.e., a map from $\mathcal{P}$ to $\mathcal{L}$) that minimizes the energy function
\begin{equation}
E(l) = \sum_{p\in \mathcal{P}} D_p(l_p) + \sum_{(p,q)\in \mathcal{N}} S_{p,q}(l_p,l_q),
\label{eq_sc}
\end{equation}
where $\mathcal{N}\subset \mathcal{P}\times \mathcal{P}$ is a neighborhood system of pixels. The \emph{smoothness term} $S_{p,q}$ is defines as
\begin{align}
S_{p,q}(l_p,l_q) & = \frac{1}{2}|l_p-l_q|(I_d(p)+I_d(q))\label{eq_smooth},\\
I_d(\cdot) &  = \|I_0(\cdot)-I_1(\cdot)\|_2 \label{eq_diff},
\end{align}
where $I_d(\cdot)$ denotes the color difference map. The \emph{data term} $D_p(l_p)$ measures the penalty of assigning 
pixel $p$ with label $l_p$, we refer to \cite{li2018perception} for more details.

%The \emph{data term} $D_p$ is defined as
%\begin{equation}\label{eq_data}
%\left\{\begin{array}{ll}
%D_p(1)=0,~D_p(0)=\mu, & \mbox{ if } p\in\partial I_0\cap\partial\mathcal{P}, \\
%D_p(0)=0,~D_p(1)=\mu, & \mbox{ if } p\in\partial I_1\cap\partial\mathcal{P}, \\
%D_p(0)=D_p(1)=0, & \mbox{ otherwise,}
%\end{array}
%\right.
%\end{equation}
%where $\mu$ is a very large penalty to avoid mislabeling and 

The energy function (\ref{eq_sc}) is then minimized via graph cuts \cite{boykov2001fast} to obtain the seam.

\subsection{Patch-point Evaluation Algorithm}

\begin{figure}
	\centering
	\subfloat[]{
		\includegraphics[height=0.13\textheight]{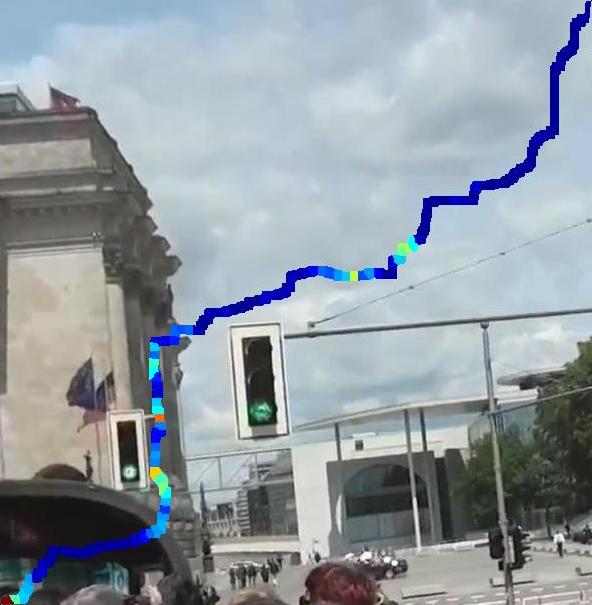}
		\includegraphics[height=0.13\textheight]{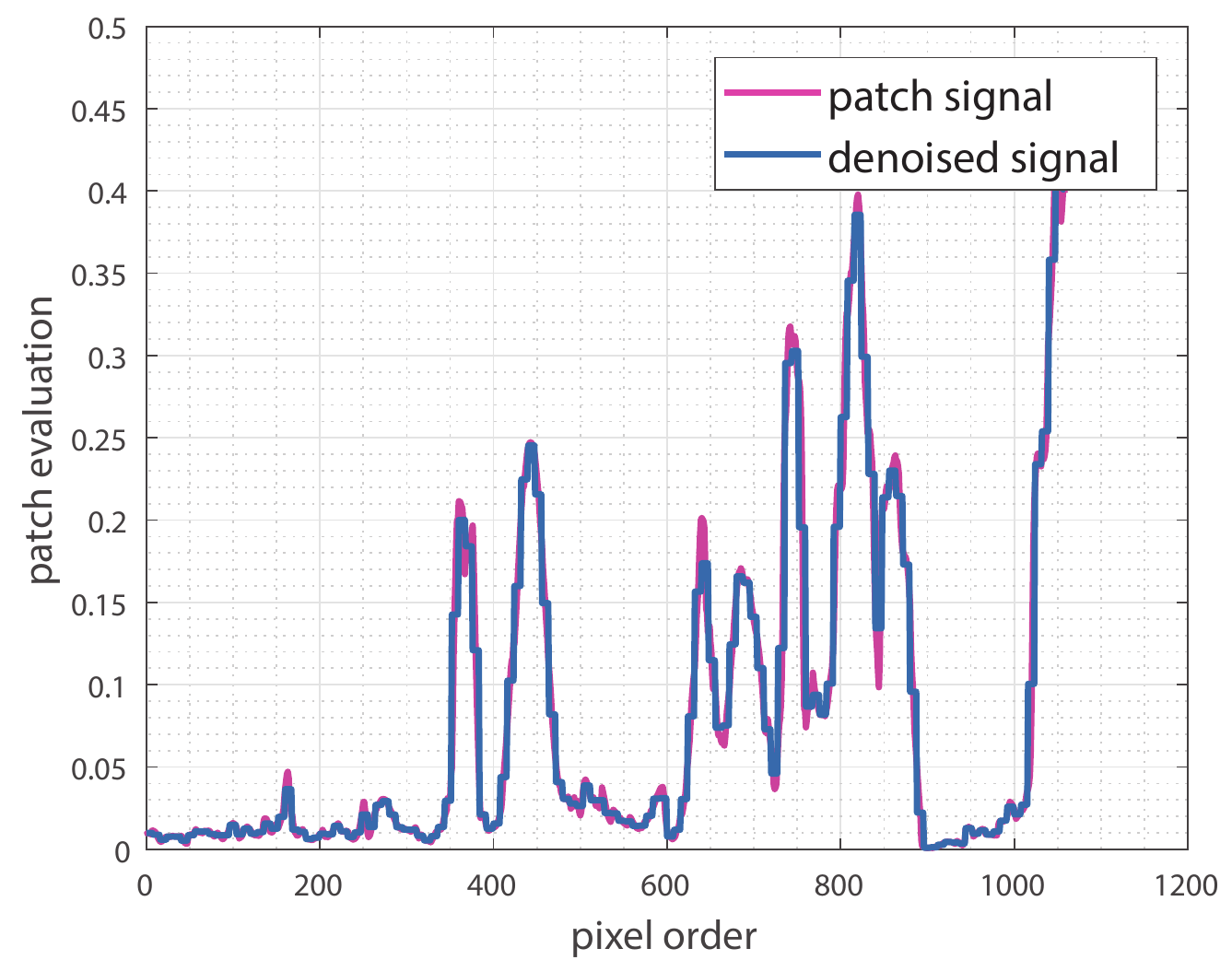}}\\	
	\subfloat[]{
		\includegraphics[height=0.13\textheight]{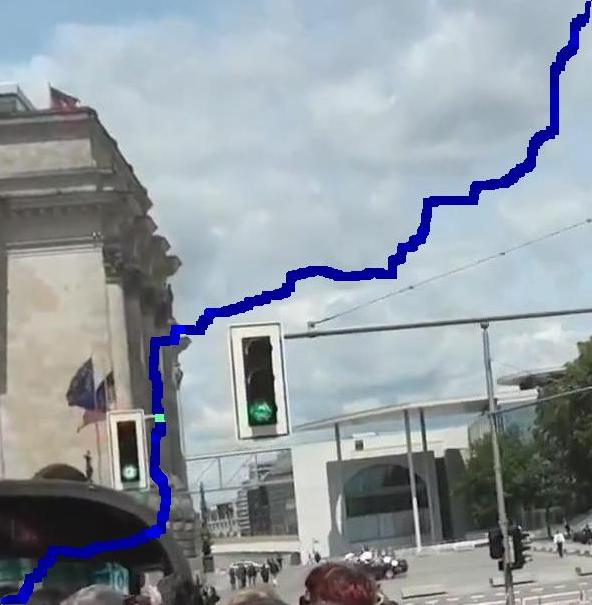}
		\includegraphics[height=0.13\textheight]{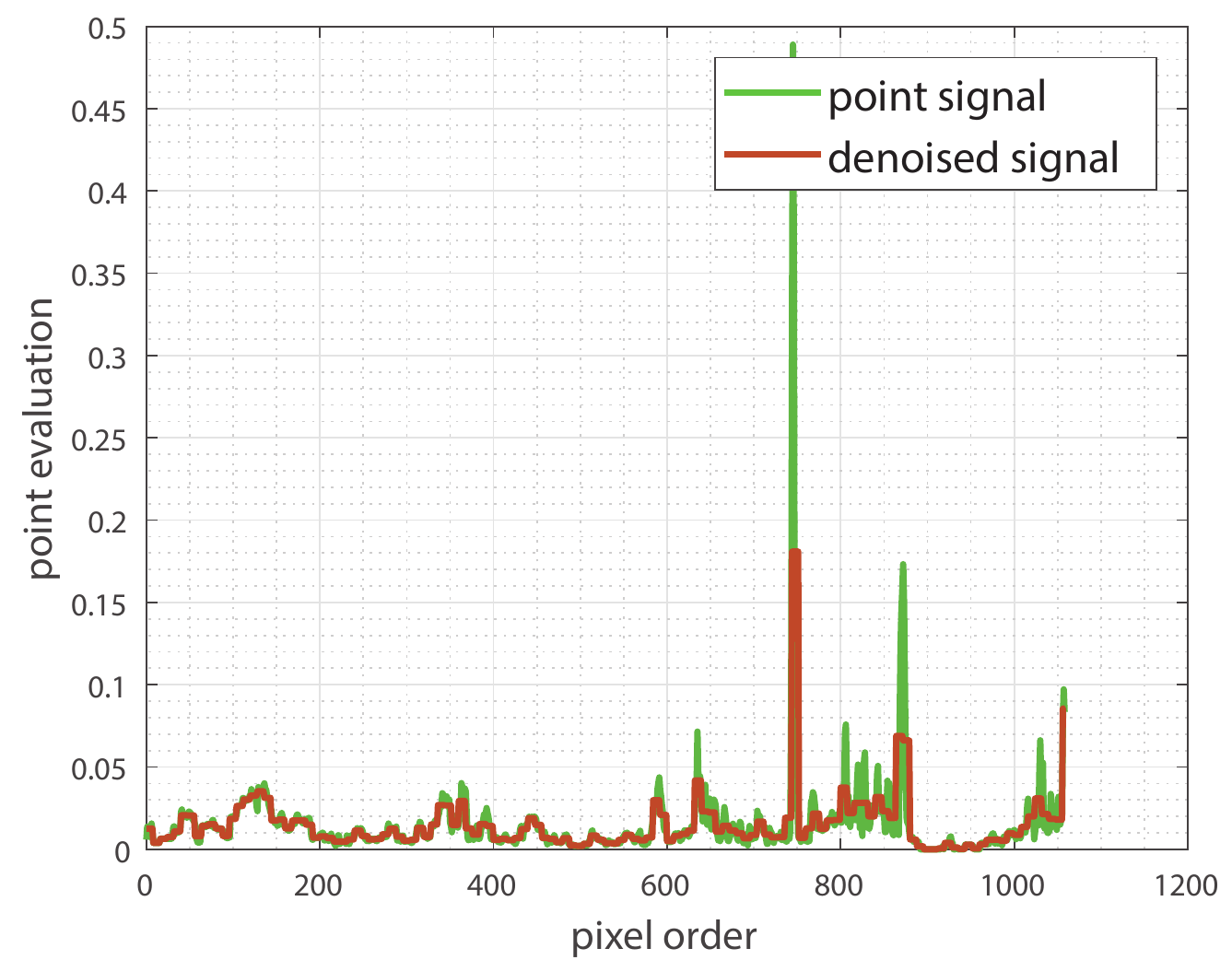}}\\
	\subfloat[]{
		\includegraphics[height=0.13\textheight]{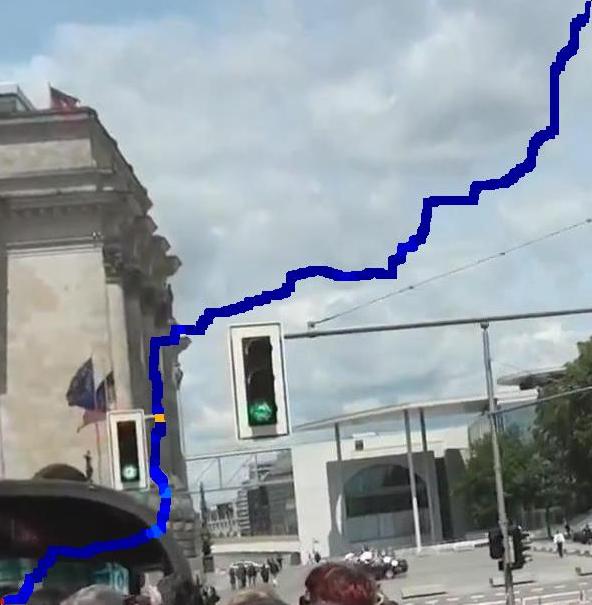}
		\includegraphics[height=0.13\textheight]{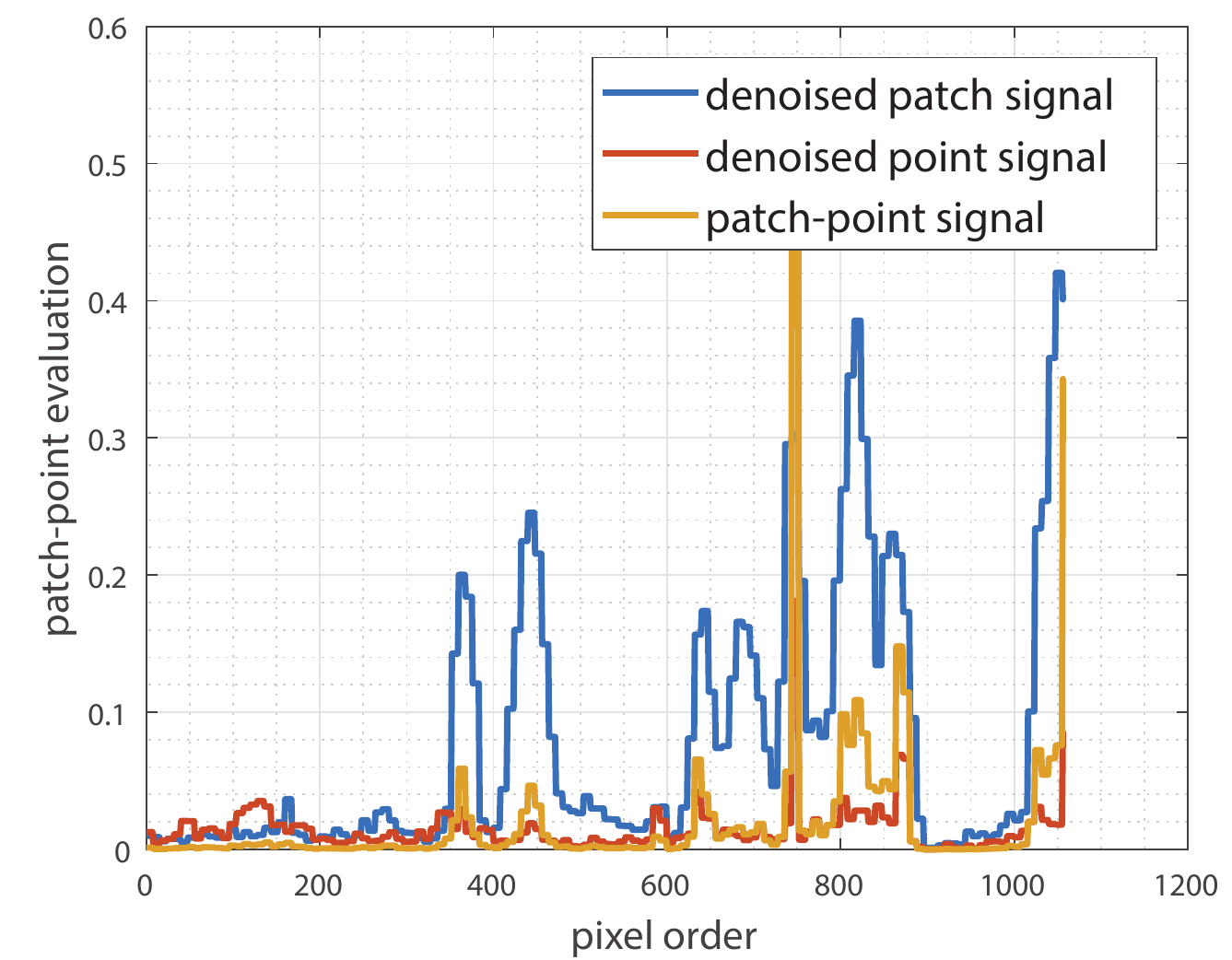}}\\
	\caption{Patch-point evaluation algorithm. Input images are from \cite{zhang2014parallax}, seams are shown as a hot map. (a) Seam and the patch signal. (b) Seam and the point signal. (c) Seam and the patch-point signal. The $x$-axis of the signals is the order of pixels along the seam, $y$-axis is the value of evaluations.}
	\label{fig:patch}
\end{figure}

\begin{figure*}
	\centering
	\subfloat[]{
		\includegraphics[width=0.24\textwidth]{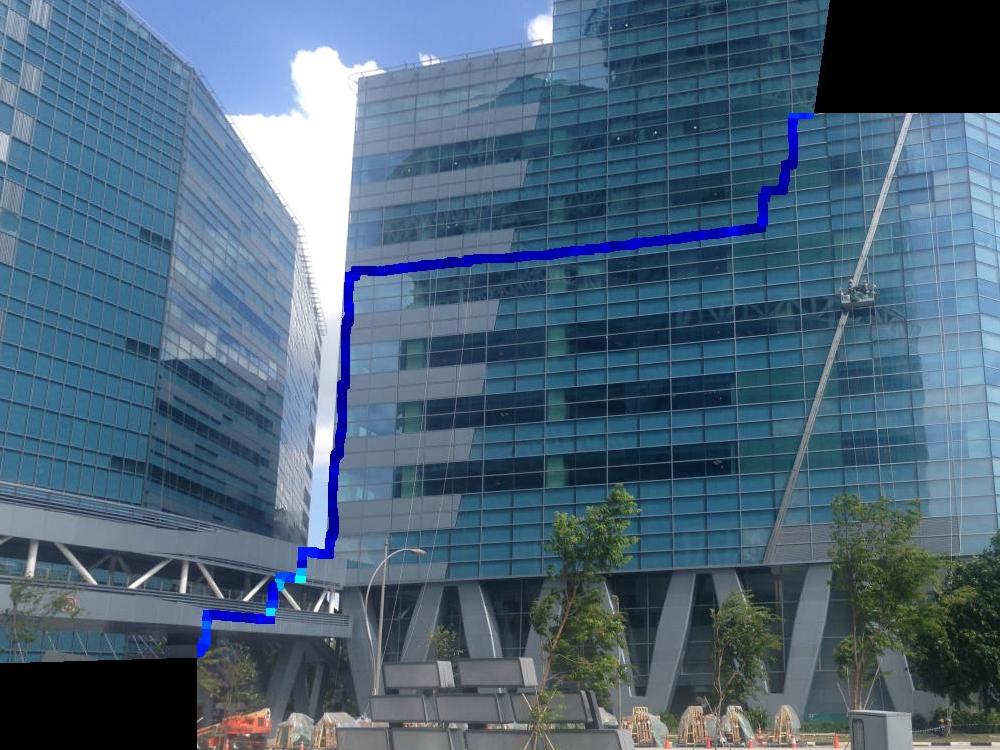}}
	\subfloat[]{
		\includegraphics[width=0.24\textwidth]{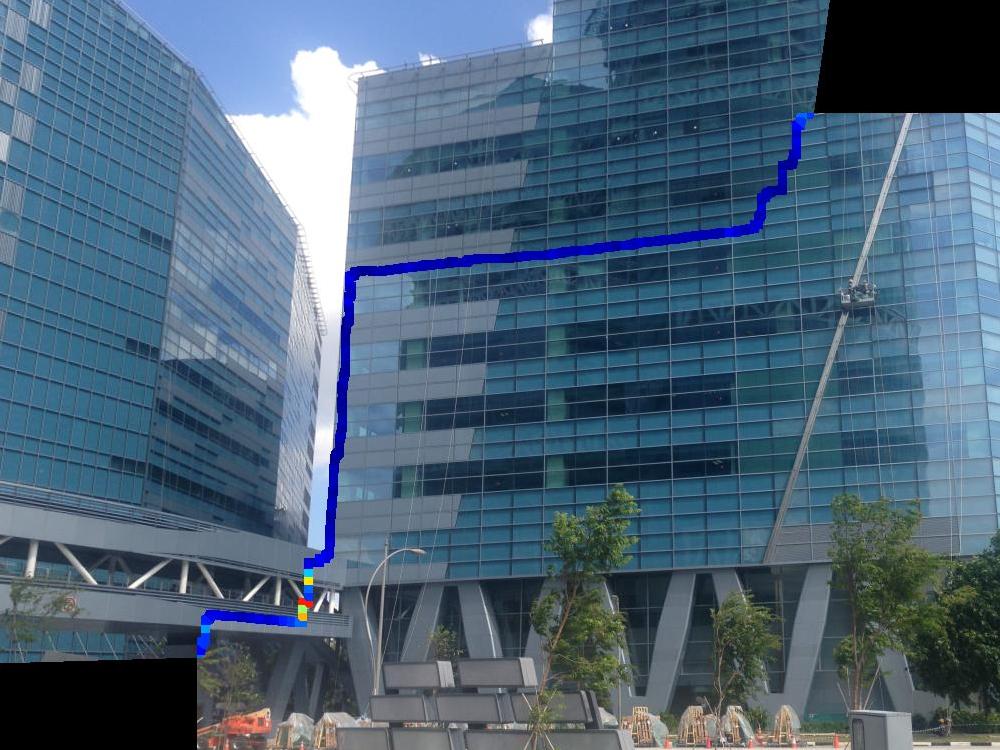}}
	\subfloat[]{
		\includegraphics[width=0.24\textwidth]{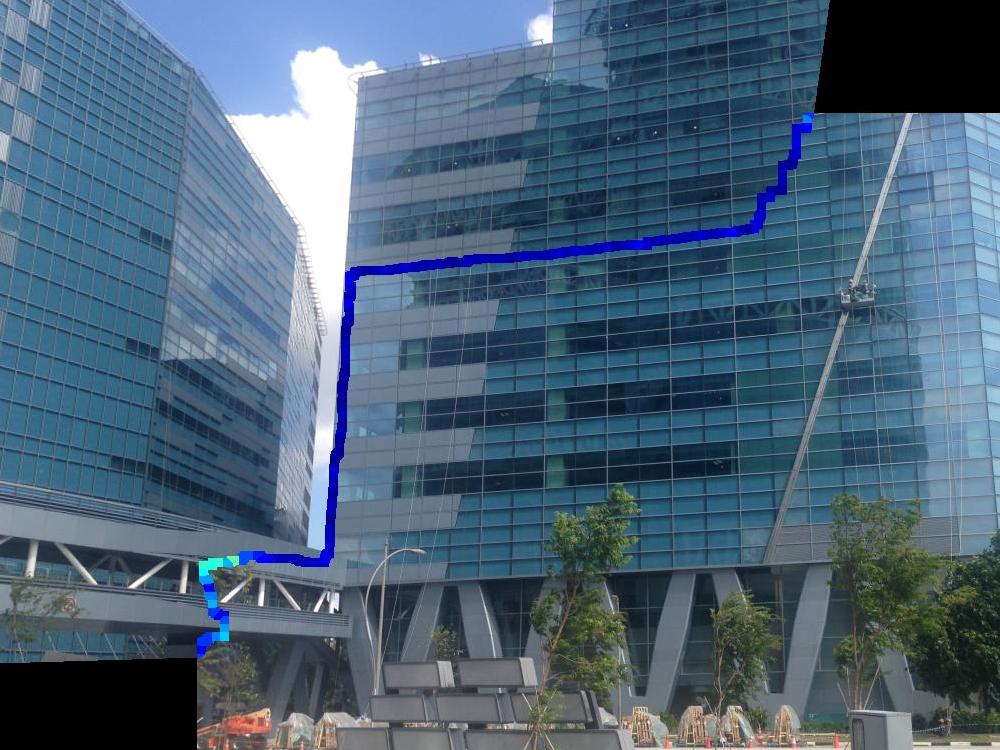}}
	\subfloat[]{
		\includegraphics[width=0.24\textwidth]{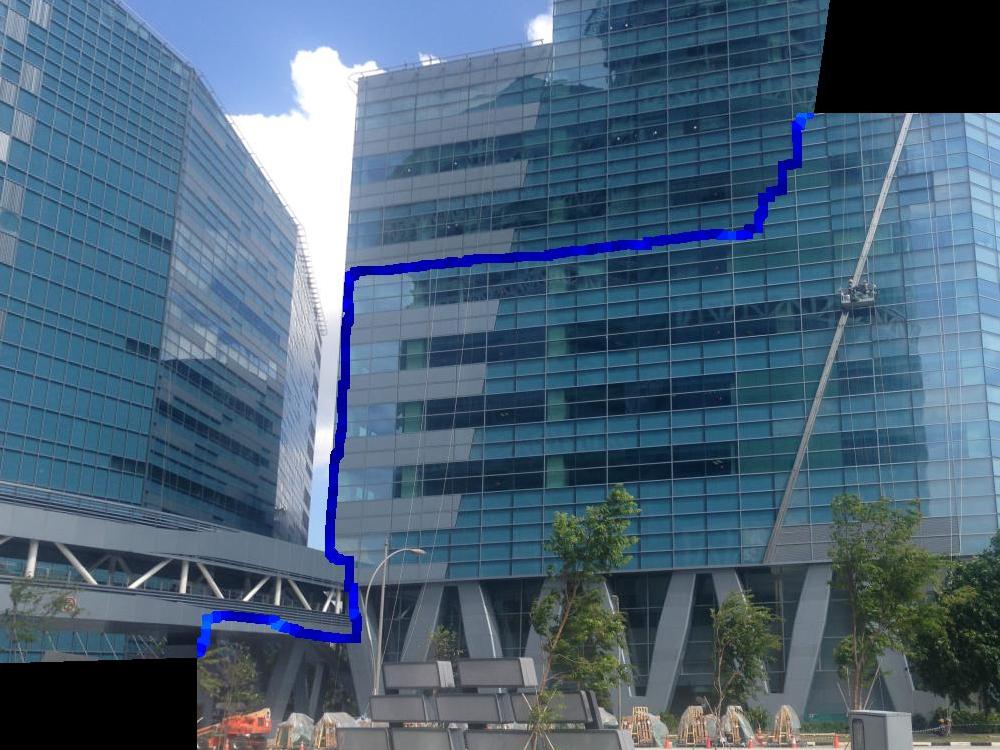}}\\
	\caption{Seam estimation refinement. Input images are from \cite{lin2016seagull}. All the results are partially cropped for the sake of layout. (a) The initial estimated seam. (b) The middle estimated seam. (c) The latter estimated seam.  (d) The final estimated seam.}
	\label{fig:refine}
\end{figure*}

To evaluate the stitching seam, a ZNCC-based method was proposed by \cite{lin2016seagull} and further used in \cite{li2018perception}. For each pixel $p_i$ on the seam, they extract a local patch centered at $p_i$ and compute the ZNCC score between the local patch in the target image $I_1$ and that in the reference image $I_0$. The seam quality is defined as
\begin{equation}\label{eq_zncc}
Q_{seam} = \frac{1}{N}\sum_i^N(\frac{1-\text{ZNCC}(p_i)}{2}),
\end{equation}
where $N$ is the total number of pixels on the seam. As shown in Fig. \ref{fig:1}, such quality measures the average performance of these pixels without considering the relevance and variance among them. It may cause that the seam with the minimal measure is not optimal in human perception.

Despite the difficulties of defining a precise seam quality metric, we can still use this strategy to evaluate the pixels on the seam. Generally, the patch differences have a good ``continuity'' property while the point differences have a nice ``diversity'' property (see Fig. \ref{fig:patch}). Thus, we combine the patch and point together to evaluate the seam.

\subsubsection{Patch evaluation}

As the misalignment artifacts usually occur as structural inconsistency in the overlapping region, we use the SSIM (structural similarity) index \cite{wang2004image} instead of ZNCC to compare the local patches in the two images. Experiments also show the superior robustness of SSIM. The patch evaluation on pixel $p_i$ is defined as
\begin{equation}
E_{patch}(p_i) = \frac{1 - \text{SSIM}(p_i)}{2}.
\label{eq_patch}
\end{equation}
The SSIM index is a decimal value between $-1$ and $1$, and value $1$ is only reachable if the two local patches are identical. Thus, a misaligned pixel on the seam usually possesses a relatively large value of patch evaluation.

\subsubsection{Point evaluation}

For parallax issues in imperfect image series, a single patch evaluation is not enough to provide a precise evaluation for the pixels on the seam. Sometimes it will create false positives, which gives some well-aligned pixels a relatively large value of patch evaluation (see Fig. \ref{fig:patch}(a)). We add a point evaluation for the pixels to improve the evaluation algorithm. The point evaluation on pixel $p_i$ is defined as
\begin{equation}
E_{point}(p_i) = \frac{\|I_0(p_i)-I_1(p_i)\|_2 + \|I_0(q_i)-I_1(q_i)\|_2}{2},
\label{eq_point}
\end{equation}
where $p_i$ and $q_i$ are adjacent in the overlapping region and $l_{p_i}\neq l_{q_i}$. The point evaluation measures the color difference between the pixels on the two sides of the seam. Thus, a plausible seam would have a relatively small value of point evaluation for (nearly) all pixels on the seam. This can avoid the false positives in the patch evaluation.

\subsubsection{Evaluation algorithm}

To investigate the correlation and variation between these pixels, we take the evaluations as signals where the $x$-axis is the order of pixels along the seam (see Fig. \ref{fig:patch}). We smooth out the signals with a wavelet denoising tool to eliminate the effect of the invisible misalignments. An alternative way is to smooth out the original aligned images via Gaussian filter, we experimentally find that the wavelet denoising way is more effective.

Generally, a misaligned pixel on the seam would simultaneously possess a large value of patch and point evaluation. We define the evaluation $E(p_i)$ for $p_i$ as follows,
\begin{equation}
E(p_i) = \lambda \cdot E_{patch}(p_i) \cdot E_{point}(p_i),
\label{eq_pp}
\end{equation}
where $\lambda$ is added to maintain the scale of the evaluation. Fig. \ref{fig:patch} shows an example of our patch-point evaluation algorithm on a stitching seam, where the evaluation $E(p_i)$ for each pixel $p_i$ is shown as a hot map. We can see that the evaluations are nearly consistent with the human perception.

\begin{figure*}
	\centering
	\includegraphics[height=0.46\textheight]{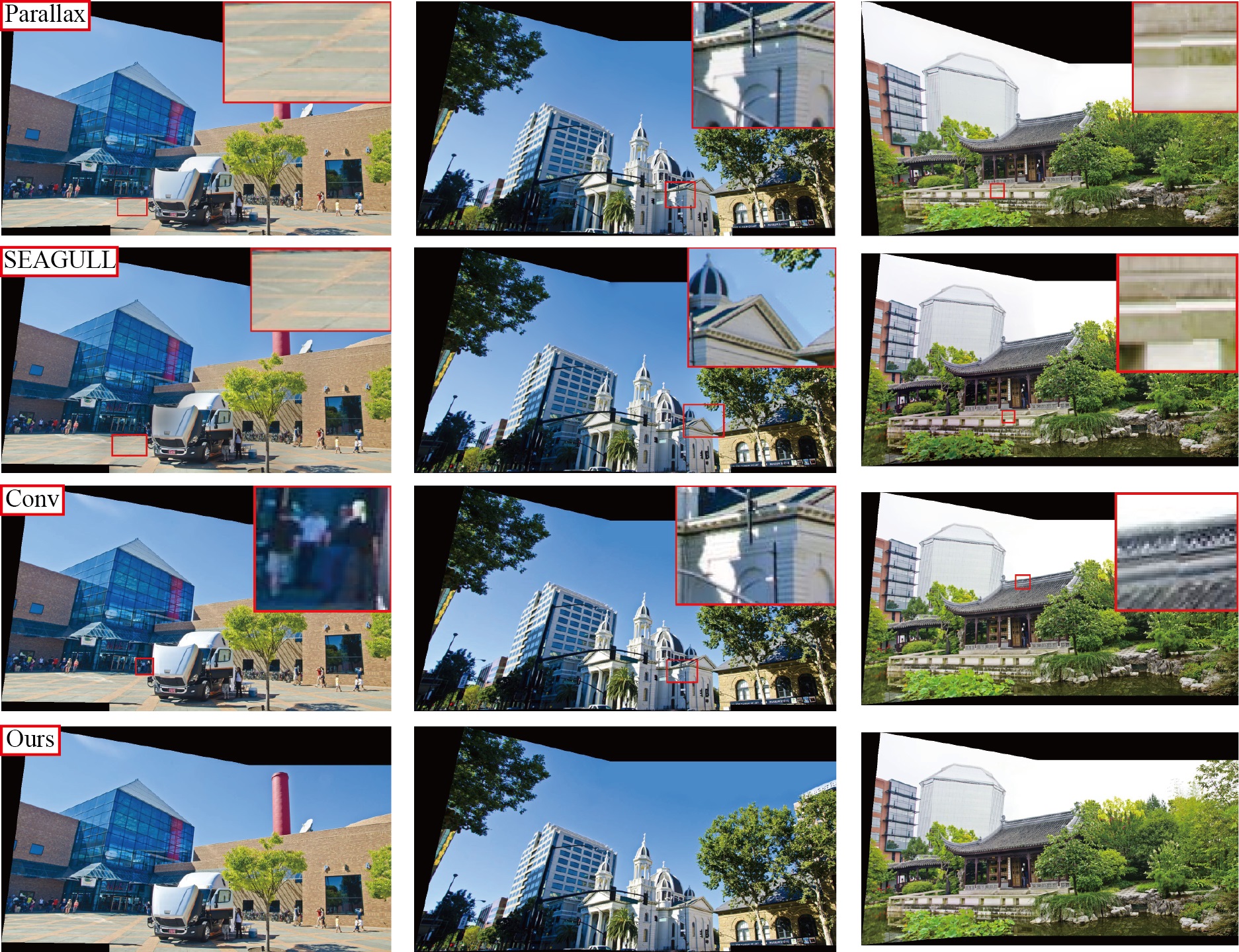}\\
	\caption{Comparisons between different stitching methods. Image datasets are from Parallax \cite{zhang2014parallax}. From top to bottom: Parallax, SEAGULL, Conventional seam-cutting (Conv) and Ours.}
	\label{fig:comparison}
\end{figure*}

\subsection{Seam Estimation Refinement}

We then utilize the patch-point evaluation to iteratively refine our seam estimation. In general, a misaligned pixel on the seam would possess a large value of patch-point evaluation, on the contrary, a relatively large value of the patch-point evaluation usually means a potential misaligned pixel. Thus, in the seam refinement, we increase the smoothness costs of the potential misaligned pixels by modifying the difference map with
\begin{equation}
f(x) = e^{\sigma(x-\epsilon)}.
\label{eq_fx}
\end{equation}
Then, the difference map of the overlapping region is turning into
\begin{equation}
\tilde{I}_d(p) = \left\{\begin{array}{ll}
f(E(p)) \cdot I_d(p),  & \mbox{$p\in \mathcal{N}(\text{seam})$},\\
I_d(p),   & \mbox{otherwise.}
\end{array}\right.
\label{eq_diff_refine}
\end{equation}
We use $\mathcal{N}(\text{seam})$ to denote a banding area containing the seam, which is generated by expanding the seam for 5 pixels on its each side. For pixel $p\in \mathcal{N}(\text{seam})$, $E(p)$ is set to be the patch-point evaluation of its nearest pixel on the seam. The difference map is recalculated in the banding area for efficiency and robustness.

We then recalculate the energy function with the new difference map and reestimate a stitching seam. The evaluation-reestimation procedure iterates until the current seam changes negligibly comparing with the previous seams. Here ``negligibly'' means that the current seam can be totally contained in the previous banding areas. For a reasonable initial seam, this procedure usually terminates within 5 iterations. Finally, we obtain a stitching seam and the final result is generated by applying the gradient domain fusion \cite{Perez:2003} on the seam.

Fig. \ref{fig:refine} shows a stitching example of the seam estimation refinement where the seam in each iteration is shown as a hot map. The initial estimated seam suffers from the artifacts of structural inconsistency as it passes through the misaligned regions. With several iterations, we can obtain a perceptually convincing seam.

\subsection{Proposed Coarse-to-fine Framework}

We summarize our coarse-to-fine seam estimation framework in Algorithm \ref{algor:1}.

\begin{algorithm}
	\caption{Coarse-to-fine seam estimation.}
	\label{algor:1}
	\begin{algorithmic}[1]
		\REQUIRE two aligned images $I_0$ and $I_1$;
		\ENSURE a final stitching seam $S_*$;
		\STATE initial: banding area $\mathcal{B}=\emptyset$;
		\STATE calculate a color difference map $I_d$ via (\ref{eq_diff});
		\STATE calculate Eq. (\ref{eq_smooth}) and minimize Eq. (\ref{eq_sc}) via graph cuts \cite{boykov2001fast} to obtain a stitching seam $S_1$;
		\WHILE {$S_1\not\subset  \mathcal{B}$}
		\STATE conduct the patch-point evaluation algorithm on $S_1$ with Eq. (\ref{eq_patch},\ref{eq_point},\ref{eq_pp});
		\STATE expand $S_1$ to a banding area $\mathcal{N}(S_1)$ and recalculate a color difference map $\tilde{I}_d$ with Eq. (\ref{eq_fx},\ref{eq_diff_refine});
		\STATE recalculate Eq. (\ref{eq_sc}) with (\ref{eq_smooth},\ref{eq_diff_refine}) and reestimate a stitching seam $S_2$;
		\STATE set $\mathcal{B} = \mathcal{B}\cup \mathcal{N}(S_1)$, let $I_d = \tilde{I}_d$ and $S_1 = S_2$;
		\ENDWHILE
		\RETURN $S_* = S_1$.
	\end{algorithmic}
\end{algorithm}

\section{Experiments}

In our experiments, the patch size in the patch evaluation is set to be $21\times 21$, $\lambda$ in (\ref{eq_pp}) equals $10$, $\sigma$ and $\epsilon$ in (\ref{eq_fx}) are set to be $5$ and $0.12$ respectively. We use SIFT \cite{lowe2004distinctive} and RANSAC \cite{fischler1981random} to find the feature correspondences between input images. Global homography or other available warps \cite{zaragoza2014projective,lin2016seagull} are then estimated to align the images. Finally, our coarse-to-fine seam estimation method is adopted to estimate the final seam and the final result is generated by blending the aligned images via gradient domain fusion \cite{Perez:2003}.

We compare our method with the conventional seam-cutting and other seam-driven methods \cite{zhang2014parallax,lin2016seagull}. The comparisons are done on public available datasets including Parallax \cite{zhang2014parallax} and SEAGULL \cite{lin2016seagull}. All comparison results are provided in the supplementary material.

Fig. \ref{fig:comparison} shows some comparisons between different stitching methods. Input images are from \cite{zhang2014parallax}. The conventional seam-cutting, SEAGULL and our method adopt the same image alignment provided by SEAGULL. Parallax, SEAGULL and conventional seam-cutting suffer from the visual artifacts of structural inconsistency, as shown in red rectangle. Our method can finally produce convincing results in human perception.

\section{Conclusion}

In this paper, we propose a coarse-to-fine seam estimation method to handle the imperfect image series in image stitching. Comprehensive experiments demonstrate that our method can finally find a nearly perception-consistent stitching seam after several iterations, which outperforms the conventional seam-cutting and other seam-driven methods. 
%In the future, we plan to design a seam evaluation algorithm to objectively and precisely measure the stitching seam.

%\appendices
%% you can choose not to have a title for an appendix
%% if you want by leaving the argument blank
%\section{}
%% use section* for acknowledgment
%\section*{Acknowledgment}

\ifCLASSOPTIONcaptionsoff
  \newpage
\fi



\newpage
\vfill
\balance

%\bibliographystyle{IEEEtran}
%\bibliography{IEEEabrv,bib_submit}


%\end{IEEEbiography}
%% insert where needed to balance the two columns on the last page with
%%\newpage


%\vfill


\end{document}